\documentclass[runningheads]{llncs}

\usepackage{graphicx}
\usepackage{tikz}
\usepackage{comment}
\usepackage{amsmath,amssymb} %
\usepackage{color}
\usepackage{mathtools}
\usepackage{cite}
\usepackage{caption}
\usepackage{subcaption}
\usepackage{multirow}
\usepackage{hyperref}

\begin{document}
\newcommand{\new}[1]{\textcolor{red}{#1}}
\newcommand{\valsays}[1]{\textcolor{red}{[Val: #1]}}
\newcommand{\chensays}[1]{\textcolor{blue}{[Chen: #1]}}
\newcommand{\karsays}[1]{\textcolor{blue}{[KA: #1]}}
\newcommand{\corsays}[1]{\textcolor{cyan}{[Cordelia: #1]}}

\newcommand\inner[2]{\langle #1, #2 \rangle}
\DeclarePairedDelimiter\abs{\lvert}{\rvert}%
\DeclarePairedDelimiter\norm{\lVert}{\rVert}%

\pagestyle{headings}
\mainmatter
\def\ECCVSubNumber{1710}  %

\title{Multi-modal Transformer for Video Retrieval} %

\titlerunning{Multi-modal Transformer for Video Retrieval}
\author{Valentin Gabeur\inst{1,2},
Chen Sun\inst{2},
Karteek Alahari\inst{1},
Cordelia Schmid\inst{2}
}
\authorrunning{V.\ Gabeur et al.}
\institute{Inria\footnote[1]{Univ. Grenoble Alpes, Inria, CNRS, Grenoble INP, LJK, 38000 Grenoble, France.}, 
\email{karteek.alahari@inria.fr}\\
\and
Google Research, 
\email{\{valgab,chensun,cordelias\}@google.com}\\
}
\maketitle

\begin{abstract}
The task of retrieving video content relevant to natural language queries plays a critical role in effectively handling internet-scale datasets. Most of the existing methods for this caption-to-video retrieval problem do not fully exploit cross-modal cues present in video. 
Furthermore, they aggregate per-frame visual features with limited or no temporal information.
In this paper, we present a multi-modal transformer to jointly encode the different modalities in video, which allows each of them to attend to the others. The transformer architecture is also leveraged to encode and model the temporal information. On the natural language side, we investigate the best practices to jointly optimize the language embedding together with the multi-modal transformer. This novel framework allows us to establish state-of-the-art results for video retrieval on three datasets. More details are available at~\url{http://thoth.inrialpes.fr/research/MMT}.

\end{abstract}

\section{Introduction}
Video is one of the most popular forms of media due to its ability to capture dynamic events and its natural appeal to our visual and auditory senses. Online video platforms are playing a major role in promoting this form of media. However, the billions of hours of video available on such platforms are unusable if we cannot access them effectively, for example, by retrieving relevant content through queries.

In this paper, we tackle the tasks of caption-to-video and video-to-caption retrieval. In the first task of caption-to-video retrieval, we are given a query in the form of a caption (e.g., ``How to build a house") and the goal is to retrieve the videos best described by it (i.e., videos explaining how to build a house). In practice, given a test set of caption-video pairs, our aim is to provide, for each caption query, a ranking of all the video candidates such that the video associated with the caption query is ranked as high as possible.
On the other hand, the task of video-to-caption retrieval focuses on finding among a collection of caption candidates the ones that best describe the query video.

A common approach for the retrieval problem is similarity learning~\cite{Xing2003distancemetric}, where we learn a function of two elements (a query and a candidate) that best describes their similarity. All the candidates can then be ranked according to their similarity with the query. In order to perform this ranking, the captions as well as the videos are represented in a common multi-dimensional embedding space, wherein similarities can be computed as a dot product of their corresponding representations. The critical question here is how to learn accurate representations of both caption and video to base our similarity estimation on.

\begin{figure*}[t]
\begin{center}
\includegraphics[clip, trim={1.3cm 9cm 1.2cm 3.1cm}, width=\textwidth]{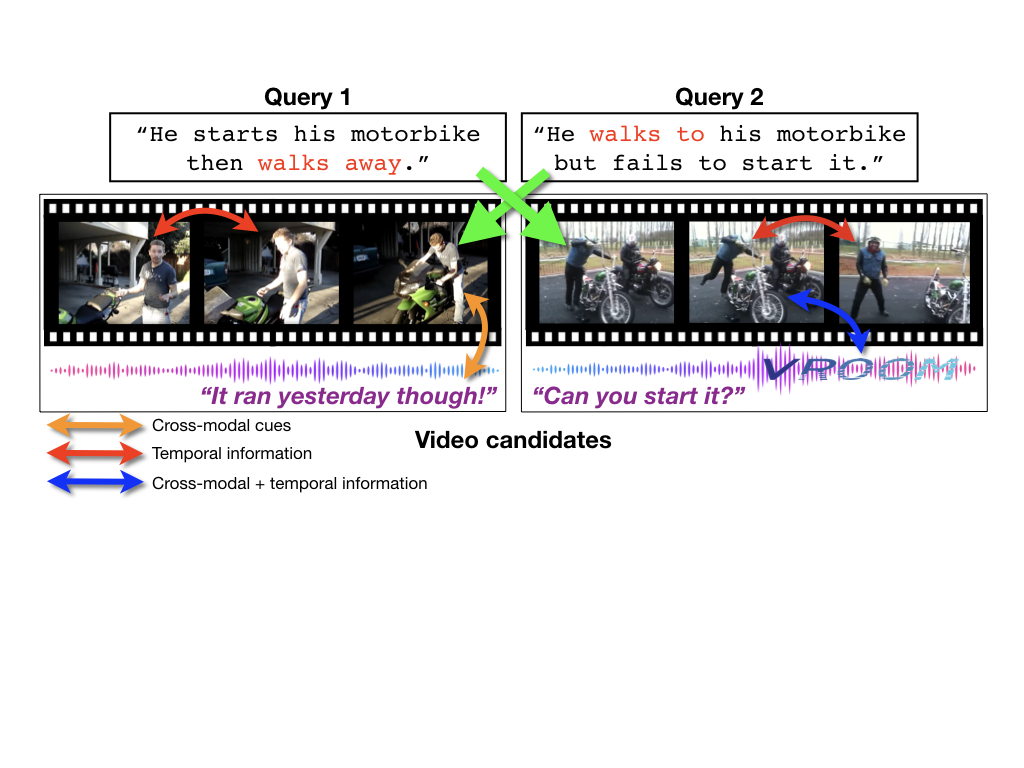}
\end{center}
   \caption{When matching a text query with videos, the inherent cross-modal and temporal information in videos needs to be leveraged effectively, for example, with a video encoder that handles all the constituent modalities (appearance, audio, speech) jointly across the entire duration of the video. In this example, a video encoder will be able to distinguish between ``someone walking \textit{to}" and ``someone walking \textit{away}" only if it exploits the temporal information of events occurring in the video (red arrows). Also, in order to understand that a ``motorbike failed to start", it needs to use cross-modal information (e.g., absence of noise after someone tried to start the engine, orange arrow).}
\label{fig:video_signal}
\end{figure*}

The problem of learning representation of text has been extensively studied, leading to various methods~\cite{Zhang2010BagOfWords,Mikolov2013Word2Vec,vaswani2017transformer,Hochreiter1997lstm,devlin2018bert}, which can be used to encode captions. In contrast to these advances, learning effective video representation continues to be a challenge, and forms the focus of our work. This is in part due to the multimodal and temporal nature of video. Video data not only varies in terms of appearance, but also in possible motion, audio, overlaid text, speech, etc. Leveraging cross-modal relations thus forms a key to building effective video representations. As illustrated in Fig.~\ref{fig:video_signal}, cues jointly extracted from all the constituent modalities are more informative than handling each modality independently. Hearing a motor sound right after seeing someone starting a bike tells us that the running bike is the visible one and not a background one. Another example is the case of a video of ``a crowd listening to a talk", neither of the modalities ``appearance" or ``audio" can fully describe the scene, but when processed together, higher level semantics can be obtained.

Recent work on video retrieval does not fully exploit such cross-modal high-level semantics. They either ignore the multi-modal signal~\cite{miech2019MIL-NCE}, treat modalities separately~\cite{miech2018learning}, or only use a gating mechanism to modulate certain modality dimensions~\cite{liu2019use}.  
Another challenge in representing video is its temporality. Due to the difficulty in handling variable duration of videos, current approaches~\cite{miech2018learning,liu2019use} discard long-term temporal information by aggregating descriptors extracted at different moments in the video. We argue that this temporal information can be important to the task of video retrieval.
As shown in Fig.~\ref{fig:video_signal}, a video of ``someone walking \textit{to} an object" and ``someone walking \textit{away} from an object" will have the same representation once pooled temporally, however, the movement of the person relative to the object is potentially important in the query.

\begin{figure*}[t]
\begin{center}
\includegraphics[clip, trim=0cm 0cm 0cm 0cm, width=1.0\textwidth]{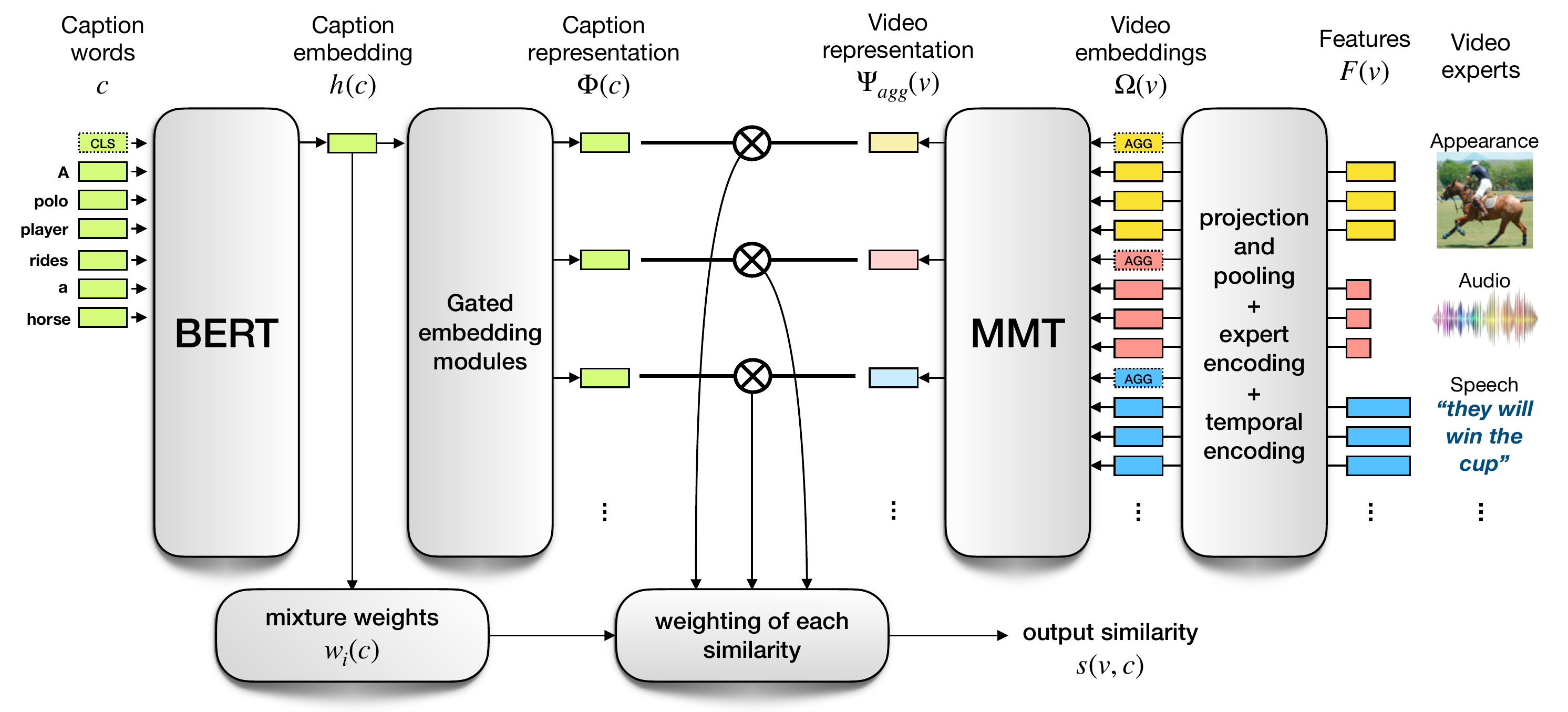}
\end{center}
   \caption{Our cross-modal framework for similarity estimation. We use our Multi-modal Transformer (MMT, right) to encode video, and BERT (left) for text.}
\label{fig:architecture}
\end{figure*}

We address the temporal and multi-modal challenges posed in video data by introducing our multi-modal transformer. 
It performs the task of processing features extracted from different modalities at different moments in video and aggregates them in a compact representation.
Building on the transformer architecture~\cite{vaswani2017transformer}, our multi-modal transformer exploits the self-attention mechanism to gather valuable cross-modal and temporal cues about events occurring in a video. We integrate our multi-modal transformer in a cross-modal framework, as illustrated in Fig.~\ref{fig:architecture}, which leverages both captions and videos, and estimates their similarity.

\paragraph{Contributions.}
 In this work, we make the following three contributions:
(i) First, we introduce a novel video encoder architecture for retrieval: Our multi-modal transformer processes effectively multiple modality features extracted at different times.
(ii) We thoroughly investigate different architectures for language embedding, and show the superiority of the BERT model for the task of video retrieval. (iii) By leveraging our novel cross-modal framework we outperform prior state of the art for the task of video retrieval on MSRVTT~\cite{xu2016msrvtt}, ActivityNet~\cite{krishna2017activitynet} and LSMDC~\cite{Rohrbach2015LSMDC} datasets. It is also the winning solution in the CVPR 2020 Video Pentathlon Challenge~\cite{Gabeur2020Challenge}.

\section{Related work}

We present previous work on language and video representation learning, as well as on visual-language retrieval.

\vspace{0.2cm}
\noindent \textbf{Language representations.} 
Earlier work on language representations include bag of words~\cite{Zhang2010BagOfWords} and Word2Vec~\cite{Mikolov2013Word2Vec}. A limitation of these representations is capturing the sequential properties in a sentence. LSTM~\cite{Hochreiter1997lstm} was one of the first successful deep learning models to handle this. More recently, the transformer~\cite{vaswani2017transformer} architecture has shown impressive results for text representation by implementing a self-attention mechanism where each word (or wordpiece~\cite{Wu2016Wordpiece}) of the sentence can attend to all the others. The transformer architecture, consisting of self-attention layers alternatively stacked with fully-connected layers, forms the base of the popular language modeling network BERT~\cite{devlin2018bert}. Burns et al.~\cite{burns2019language} perform an analysis of the different word embeddings and language models (Word2Vec~\cite{Mikolov2013Word2Vec}, LSTM~\cite{Hochreiter1997lstm}, BERT~\cite{devlin2018bert}, etc.) used in vision-language tasks.
They show that the pretrained and frozen BERT model~\cite{devlin2018bert} performs relatively poorly compared to a LSTM or even a simpler average embedding model. In this work, we show that for video retrieval, a pretrained BERT outperforms other language models, but it needs to be finetuned.

\vspace{0.2cm}
\noindent \textbf{Video representations.} With a two-stream network, Simonyan et al.~\cite{Simonyan2014TwoStream} have used complementary information from still frames and motion between frames to perform action recognition in videos. Carreira et al.~\cite{Carreira2017Conv3D} incorporated 3D convolutions in a two-stream network to better attend the temporal structure of the signal. S3D~\cite{Xie2017S3D} is an alternative approach, which replaced the expensive 3D spatio-temporal convolutions by separable 2D and 1D convolutions.
More recently, transformer-based methods, which leverage BERT pretraining~\cite{devlin2018bert}, have been applied to S3D features in VideoBERT~\cite{Sun2019VideoBERT} and CBT~\cite{Sun2019cbt}.
While these works focus on visual signals, they have not studied how to encode the other multi-modal semantics, such as audio signals.

\vspace{0.2cm}
\noindent \textbf{Visual-language retrieval.} Harwath et al.~\cite{Harwath2018} perform image and audio-caption retrieval by embedding audio segments and image regions in the same space and requiring high similarity between each audio segment and its corresponding image region. The method presented in~\cite{lee2018stacked} takes a similar approach for image-text retrieval by embedding images regions and words in a joint space. A high similarity is obtained for images that have matching words and image regions.

For videos, JSFusion~\cite{Yu2018JSFusion} estimates video-caption similarity through dense pairwise comparisons between each word of the caption and each frame of the video. In this work, we instead estimate both a video embedding and a caption embedding and then compute the similarity between them. Zhang et al.~\cite{zhang2018HSE} perform paragraph-to-video retrieval by assuming a hierarchical decomposition of the video and paragraph. Our method do not assume that the video can be decomposed into clips that align with sentences of the caption. A recent alternative is creating separate embedding spaces for different parts of speech (e.g., noun or verb)~\cite{wray2019finegrained}. In contrast to this method, we do not pre-process the sentences but encode them directly through BERT.

Another work~\cite{miech19howto100m} leverages the large number of instructional videos in the HowTo100M dataset, but does not fully exploit the temporal relations. Our work instead relies on longer segments extracted from HowTo100M videos in order to learn temporal dependencies and address the problem of misalignment between speech and visual features.
Mithun et al.~\cite{mithun2018learning,mithun2019joint} use three experts (Object, Activity and Place) to compute three corresponding text-video similarities. These experts however do not collaborate together as their respective similarities are simply summed together. A related approach~\cite{miech2018learning} uses precomputed features from experts for text to video retrieval, where the overall similarity is obtained as a weighted sum of each expert's similarity.
A recent extension~\cite{liu2019use} to this mixture of experts model uses a collaborative gating mechanism for modulating each expert feature according to the other experts. However, this collaborative gating mechanism only strengthens (or weakens) some dimensions of the input signal in a single step, and is therefore not able to capture high level inter-modality information. Our multi-modal transformer overcomes this limitation by attending to all available modalities over multiple self-attention layers.

\section{Methodology}
Our overall method relies on learning a function $s$ to compute the similarity between two elements: text and video, as shown in Fig.~\ref{fig:architecture}. We then rank all the videos (or captions) in the dataset, according to their similarity with the query caption (or video) in the case of text-to-video (or video-to-text) retrieval. In other words, given a dataset of $n$ video-caption pairs $\{(v_1,c_1), ..., (v_n,c_n)\}$, the goal of the learnt similarity function $s(v_i,c_j)$, between video $v_i$ and caption $c_j$, is to provide a high value if $i = j$, and a low one if $i \ne j$. Estimating this similarity (described in Section~\ref{section:similarity_estimation}) requires accurate representations for the video as well as the caption. Fig.~\ref{fig:architecture} shows the two parts focused on producing these representations (presented in Sections~\ref{sec:videorep} and~\ref{sec:captionrep} respectively) in our cross-modal framework.

\subsection{Video representation}
\label{sec:videorep}

The video-level representation is computed by our proposed multi-modal transformer (MMT). MMT follows the architecture of the transformer encoder presented in~\cite{vaswani2017transformer}. It consists of stacked self-attention layers and fully collected layers. MMT's input $\Omega(v)$ is a set of embeddings, all of the same dimension $d_{model}$. Each of them embeds the semantics of a feature, its modality, and the time in the video when the feature was extracted. This input is given by:
\begin{equation} \label{eq:transformer_input}
     \Omega(v) = F(v) + E(v) + T(v),
\end{equation}

In the following, we describe those three components.

\noindent\textbf{Features $F$.} 
In order to learn an effective representation from different modalities inherent in video data, we begin with video feature extractors called ``experts''~\cite{mithun2018learning,Yu2018JSFusion,miech2018learning,liu2019use}. In contrast to previous methods, we learn a joint representation leveraging both cross-modal and long-term temporal relationships among the experts.
We use $N$ pretrained experts $\{F^n\}_{n=1}^N$. Each expert is a model trained for a particular task that is then used to extract features from video. For a video $v$, each expert extracts a sequence $F^n(v) = [F^n_1, ..., F^n_K]$ of $K$ features.

The features extracted by our experts encode the semantics of the video. Each expert $F^n$ outputs features in $\mathbb{R}^{d_n}$.
In order to project the different expert features into a common dimension $d_{model}$, we learn $N$ linear layers (one per expert) to project all the features into $\mathbb{R}^{d_{model}}$.

A transformer encoder produces an embedding for each of its feature inputs, resulting in several embeddings for an expert. In order to obtain a unique embedding for each expert, we define an aggregated embedding $F^n_{agg}$ that will collect and contextualize the expert's information. We initialize this embedding with a max pooling aggregation of all the corresponding expert's features as $F^n_{agg} = maxpool(\{F^n_k\}_{k=1}^K)$. The sequence of input features to our video encoder then takes the form:
\begin{equation} \label{eq:features}
     F(v) = [F^1_{agg}, F^1_1, ..., F^1_K, ..., F^N_{agg}, F^N_1, ..., F^N_K].
\end{equation}

\begin{figure*}[t]
\begin{center}
\includegraphics[clip, trim=1.1cm 10cm 1cm 8cm, width=1.0\textwidth]{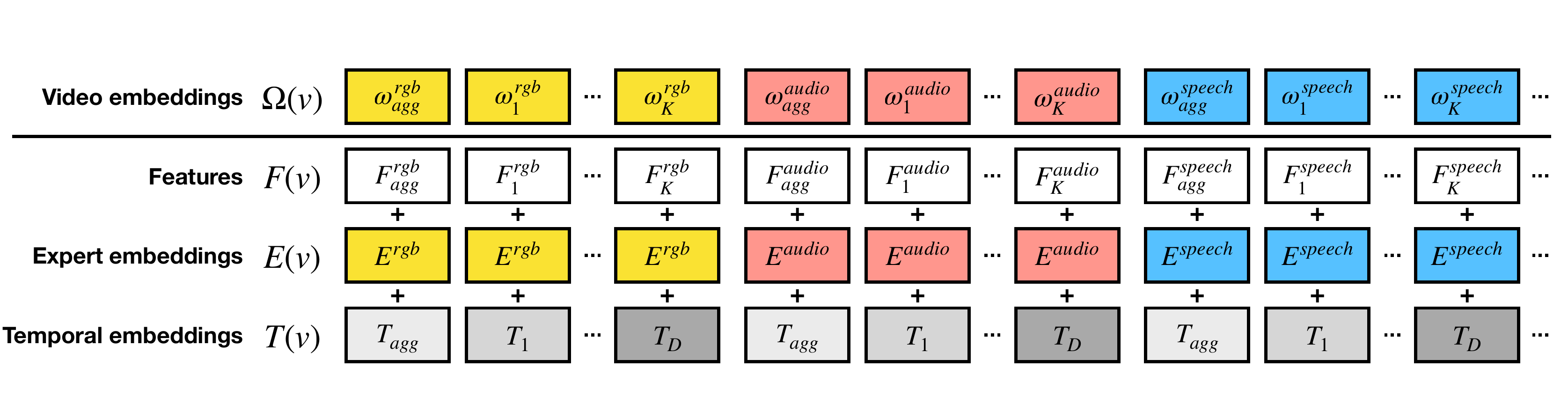}
\end{center}
   \caption{Inputs to our multi-modal transformer. We combine feature semantics $F$, expert information $E$, and temporal cues $T$ to form our video embeddings $\Omega(v)$, which are input to MMT.}
\label{fig:vid_transformer_input}
\end{figure*}

\noindent\textbf{Expert embeddings $E$.}
In order to process cross-modality information, our MMT needs to identify which expert it is attending to. We learn $N$ embeddings $\{E_1, ..., E_N\}$ of dimension $d_{model}$ to distinguish between embeddings of different experts. Thus, the sequence of expert embeddings to our video encoder takes the form:
\begin{equation} \label{eq:expert_embeddings}
     E(v) = [E^1, E^1, ..., E^1, ..., E^N, E^N, ..., E^N].
\end{equation}

\noindent\textbf{Temporal embeddings $T$.} They provide temporal information about the time in the video where each feature was extracted to our multi-modal transformer. Considering videos of a maximum duration of $t_{max}$ seconds, we learn $D = \abs{t_{max}}$ embeddings $\{T_1, ..., T_D\}$ of dimension $d_{model}$. Each expert feature that has been extracted in the time range $[t,t+1)$ will be temporally embedded with $T_{t+1}$. For example, a feature extracted at 7.4s in the video will be temporally encoded with temporal embedding $T_8$. We learn two additional temporal embeddings $T_{agg}$ and $T_{unk}$, which encode aggregated features and unknown temporal information features (for experts whose temporal information is unknown), respectively. The sequence of temporal embeddings of our video encoder then takes the form:
\begin{equation} \label{eq:temporal_embeddings}
     T(v) = [T_{agg}, T_1, ..., T_D, ..., T_{agg}, T_1, ..., T_D].
\end{equation}

\noindent\textbf{Multi-modal Transformer.}
The video embeddings $\Omega(v)$ defined as the sum of features, expert and temporal embeddings in (\ref{eq:transformer_input}), as shown in Fig.~\ref{fig:vid_transformer_input}, are input to the transformer. They are given by: $\Omega(v) = F(v) + E(v) + T(v) = [\omega^1_{agg}, \omega^1_1, ..., \omega^1_K, ..., \omega^N_{agg}, \omega^N_1, ..., \omega^N_K].$ MMT contextualises its input $\Omega(v)$ and produces the video representation $\Psi_{agg}(v)$. As illustrated in Fig.~\ref{fig:architecture}, we only keep the aggregated embedding per expert. Thus, our video representation $\Psi_{agg}(v)$ consists of the output embeddings corresponding to the aggregated features, i.e.,
\begin{equation} \label{eq:MMT processing}
     \Psi_{agg}(v) = MMT(\Omega(v))
             = [\psi^1_{agg}, ..., \psi^N_{agg}].
\end{equation}

The advantage of our MMT over the state-of-the-art collaborative gating mechanism~\cite{liu2019use} is two-fold: First, the input embeddings are not simply modulated in a single step but iteratively refined through several layers featuring multiple attention heads. Second, we do not limit our video encoder with a temporally aggregated feature for each expert, but provide all the extracted features instead, along with a temporal encoding describing at what moment of the video they were extracted from. Thanks to its self-attention modules, each layer of our multi-modal transformer is able to attend to all its input embeddings, thus extracting semantics of events occurring in the video over several modalities.

\subsection{Caption representation}
\label{sec:captionrep}
We compute our caption representation $\Phi(c)$ in two stages: first, we obtain an embedding $h(c)$ of the caption, and then project it with a function $g$ into $N$ different spaces as $\Phi = g \circ h$. For the embedding function $h$, we use a pretrained BERT model~\cite{devlin2018bert}. Specifically, we extract our single caption embedding $h(c)$ from the [CLS] output of BERT. In order to match the size of this caption representation with that of video, we learn for function $g$ as many gated embedding modules~\cite{miech2018learning} as there are video experts. Our caption representation then consists of $N$ embeddings, represented by $\Phi(c) = \{\phi^i\}_{i=1}^N$.

\subsection{Similarity estimation}
\label{section:similarity_estimation}
We compute our final video-caption similarity $s$, as a weighted sum of each expert $i$'s video-caption similarity $\inner{\phi^i}{\psi^i_{agg}}$. It is given by:
\begin{equation} \label{eq:cosine_similarity}
     s(v,c) = \sum_{i = 1}^{N} w_i(c)\inner{\phi^i}{\psi^i_{agg}},
\end{equation}
where $w_i(c)$ represents the weight for the $i$th expert. To obtain these mixture weights, we follow~\cite{miech2018learning} and process our caption representation $h(c)$ through a linear layer and then perform a softmax operation, i.e.,
\begin{equation} \label{eq:weights}
     w_i(c) = \frac{e^{h(c)^{\top} a_{i}}}{\sum_{j=1}^{N} e^{h(c)^{\top} a_{j}}},
\end{equation}
where $(a_1, ..., a_N)$ are the weights of the linear layer. The intuition behind using a weighted sum is that a caption may not describe all the inherent modalities in video uniformly. For example, in the case of a video with a person in a red dress singing opera, the caption ``a person in a red dress'' provides no information relevant for audio. On the contrary, the caption ``someone is singing'' should focus on the audio modality for computing similarity. Note that $w_i(c), \phi^i \ \text{and} \ \psi^i_{agg}$ can all be precomputed offline for each caption and for each video, and therefore the retrieval operation only involves dot product operations.

\subsection{Training}
We train our model with the bi-directional max-margin ranking loss~\cite{Karpathy2014DeepFE}:
\begin{equation} \label{eq:loss}
\mathcal{L} = \frac{1}{B}\sum_{i=1}^{B} \sum_{j \neq i} \Big[ \max(0, s_{ij} - s_{ii} + m) + \max(0, s_{ji} - s_{ii} + m)\Big],
\end{equation}
where $B$ is the batch size, $s_{ij} = s(v_{i},c_{j})$, the similarity score between video $v_i$ and caption $c_j$, and $m$ is the margin. This loss enforces the similarity for true video-caption pairs $s_{ii}$ to be higher than the similarity of negative samples $s_{ij}$ or $s_{ji}$, for all $i \neq j$, by at least $m$.

\section{Experiments}
\label{section:experiments}

\subsection{Datasets and Metrics}
\noindent\textbf{HowTo100M~\cite{miech19howto100m}}. It is composed of more than 1 million YouTube instructional videos, along with automatically-extracted speech transcriptions, which form the captions. These captions are naturally noisy, and often do not describe the visual content accurately or are temporally misaligned with it. We use this dataset only for pre-training.

\noindent\textbf{MSRVTT~\cite{xu2016msrvtt}.} This dataset is composed of 10K YouTube videos, collected using 257 queries from a commercial video search engine. Each video is 10 to 30s long, and is paired with 20 natural sentences describing it, obtained from Amazon Mechanical Turk workers. We use this dataset for training from scratch and also for fine-tuning. We report results on the train/test splits introduced in~\cite{Yu2018JSFusion} that uses 9000 videos for training and 1000 for test. We refer to this split as ``1k-A''. We also report results on the train/test split in~\cite{miech2018learning} that we refer to as ``1k-B''. Unless otherwise specified, our MSRVTT results are with ``1k-A''.

\noindent\textbf{ActivityNet Captions~\cite{krishna2017activitynet}.} It consists of 20K YouTube videos temporally annotated with sentence descriptions. We follow the approach of~\cite{zhang2018HSE}, where all the descriptions of a video are concatenated to form a paragraph. The training set has 10009 videos. We evaluate our video-paragraph retrieval on the ``val1'' split (4917 videos). We use ActivityNet for training from scratch and fine-tuning.

\noindent\textbf{LSMDC~\cite{Rohrbach2015LSMDC}.} It contains 118,081 short video clips ($\sim$ 4–5s) extracted from 202 movies. Each clip is annotated with a caption, extracted from either the movie script or the audio description. The test set is composed of 1000 videos, from movies not present in the training set. We use LSMDC for training from scratch and also fine-tuning.

\noindent\textbf{Metrics.} We evaluate the performance of our model with standard retrieval metrics: recall at rank $N$ (R@$N$, higher is better), median rank (MdR, lower is better) and mean rank (MnR, lower is better). For each metric, we report the mean and the standard deviation over experiments with 3 random seeds. In the main paper, we only report recall@5, median and mean ranks, and refer the reader to the supplementary material for additional metrics.

\subsection{Implementation details}
\label{section:experts}

\noindent\textbf{Pre-trained experts.} Recall that our video encoder uses pre-trained experts models for extracting features from each video modality. We use the following seven experts. \textbf{Motion} features are extracted from S3D~\cite{Xie2017S3D} trained on the Kinetics action recognition dataset. \textbf{Audio} features are extracted using VGGish model~\cite{Hershey2017VGGish} trained on YT8M. \textbf{Scene} embeddings are extracted from DenseNet-161~\cite{Huang2016DenselyCC} trained for image classification on the Places365 dataset~\cite{Zhou2018PlacesA1}. \textbf{OCR} features are obtained in three stages. Overlaid text is first detected using the pixel link text detection model. The detected boxes are then passed through a text recognition model trained on the Synth90K dataset. Finally, each character sequence is encoded with word2vec~\cite{Mikolov2013Word2Vec} embeddings. \textbf{Face} features are extracted in two stages. An SSD face detector is used to extract bounding boxes, which are then passed through a ResNet50 trained for face classification on the VGGFace2 dataset. \textbf{Speech} transcripts are extracted using the Google Cloud Speech to Text API, with the language set to English. The detected words are then encoded with word2vec.
\textbf{Appearance} features are extracted from the final global average pooling layer of SENet-154~\cite{Hu2017SeNet} trained for classification on ImageNet. For scene, OCR, face, speech and appearance, we use the features publicly released by~\cite{liu2019use}, and compute the other features ourselves.\\

\noindent\textbf{Training.}
For each dataset, we run a grid search on the corresponding validation set to estimate the hyperparameters. We use the Adam optimizer for all our experiments, and set the margin of the bidirectional max-margin ranking loss to 0.05. We also freeze our pre-trained expert models.

When pre-training on HowTo100M, we use a batch size of 64 video-caption pairs, an initial learning rate of 5e-5, which we decay by a multiplicative factor 0.98 every 10K optimisation steps, and train for 2 million steps. Given the long duration of most of the HowTo100M videos, we randomly sample 100 consecutive words in the caption, and keep 100 consecutive seconds of video data, closest in time to the selected words.

When training from scratch or finetuning on MSRVTT or LSMDC, we use a batch size of 32 video-caption pairs, an initial learning rate of 5e-5, which we decay by a multiplicative factor 0.95 every 1K optimisation steps. We train for 50K steps. We use the same settings when training from scratch or finetuning on ActivityNet, except for 0.90 as the multiplicative factor. 

To compute our caption representation $h(c)$, we use the ``BERT-base-cased'' checkpoint of the BERT model and finetune it with a dropout probability of 10\%. To compute our video representation $\Psi_{agg}(v)$, we use MMT with 4 layers and 4 attention heads, a dropout probability of 10\%, a hidden size $d_{model}$ of 512, and an intermediate size of 3072.

For datasets with short videos (MSRVTT and LSMDC), we use all the 7 experts and limit video input to 30 features per expert, and BERT input to the first 30 wordpieces. 
For datasets containing longer videos (HowTo100M and ActivityNet), we only use motion and audio experts, and limit our video input to 100 features per expert and our BERT input to the first 100 wordpieces.
In cases where an expert is unavailable for a given video, e.g., no speech was detected, we set the aggregated feature $F_{agg}^n$ to a zero vector. We refer the reader to the supplementary material for a study of the model complexity.

\subsection{Ablation studies and comparisons}
We will first show the advantage of pretraining our model on a large-scale, uncurated dataset. We will then perform ablations on the architecture used for our language and video encoders. Finally, we will present the relative importance of the  pretrained experts used in this work, and compare with related methods.

\noindent\textbf{Pretraining.} Table~\ref{table:remove_stop_words} shows the advantage of pretraining on HowTo100M, before finetuning on the target dataset (MSRVTT in this case). 
We also evaluated the impact of pretraining on ActivityNet and LSMDC; see Table~\ref{table:ANet_results} and Table~\ref{table:LSMDC_results}.

\begin{table}[t]
\begin{center}
\caption{Advantage of pretraining on HowTo100M then finetuning on MSRVTT. Impact of removing the stop words. Performance reported on MSRVTT.}
\label{table:remove_stop_words}
\scriptsize
\begin{tabular}{l | c | @{\hskip -0.35cm}c @{\hskip -0.35cm}c @{\hskip -0.35cm}c}
\hline
\multicolumn{2}{c}{} & \multicolumn{3}{c}{\textit{Text $\longrightarrow$ Video}}\\
Method & Caption & R@5$\uparrow$ & MdR$\downarrow$ & MnR$\downarrow$\\
\hline
\multirow{2}{*}{Pretraining without finetuning (zero-shot setting)} & all words & $\hspace{1.5em}6.9$ & $\hspace{1.5em}160.0$ & $\hspace{1.5em}240.2$ \\
 & w/o stop words & $\hspace{1.5em}\textbf{14.4}$ & $\hspace{1.5em}\textbf{66.0}$ & $\hspace{1.5em}\textbf{148.1}$ \\
\hline

\multirow{2}{*}{Training from scratch on MSRVTT} & all words & $\hspace{1.5em}\textbf{54.0}_{\!\pm\!0.2}$ & $\hspace{1.5em}\textbf{4.0}_{\!\pm\!0.0}$ & $\hspace{1.5em}\textbf{26.7}_{\!\pm\!0.9}$ \\
 & w/o stop words & $\hspace{1.5em}50.0_{\!\pm\!0.6}$ & $\hspace{1.5em}5.3_{\!\pm\!0.5}$ & $\hspace{1.5em}28.5_{\!\pm\!0.9}$ \\
\hline
\multirow{2}{*}{Pretraining then finetuning on MSRVTT} & all words & $\hspace{1.5em}\textbf{57.1}_{\!\pm\!1.0}$ & $\hspace{1.5em}\textbf{4.0}_{\!\pm\!0.0}$ & $\hspace{1.5em}\textbf{24.0}_{\!\pm\!0.8}$ \\
 & w/o stop words & $\hspace{1.5em}55.0_{\!\pm\!0.7}$ & $\hspace{1.5em}4.3_{\!\pm\!0.5}$ & $\hspace{1.5em}24.3_{\!\pm\!0.3}$ \\
\hline
\end{tabular}
\end{center}
\end{table}

\noindent\textbf{Language encoder.} We evaluated several architectures for caption representation, as shown in Table~\ref{table:language_model}. Similar to the observation made in~\cite{burns2019language}, we obtain poor results from a frozen, pretrained BERT. Using the [CLS] output from a pretrained and frozen BERT model is in fact the worst result. We suppose this is because the output was not trained for caption representation, but for a very different task: next sentence prediction. Finetuning BERT greatly improves performance; it is the best result. We also compare with GrOVLE~\cite{burns2019language} embeddings, frozen or finetuned, aggregated with a max-pooling operation or a 1-layer LSTM and a fully-connected layer. We show that  pretrained BERT embeddings aggregated by a max-pooling operation perform better than GrOVLE embeddings processed by a LSTM (best results from~\cite{burns2019language} for the text-to-clip task).

\begin{table}[t]
\begin{center}
\caption{Comparison of different architectures for caption embedding when training from scratch on MSRVTT.}
\label{table:language_model}
\scriptsize
\begin{tabular}{l | l | l| @{\hskip -0.35cm}c @{\hskip -0.35cm}c @{\hskip -0.35cm}c}
\hline
\multicolumn{2}{c}{} & \multicolumn{1}{c}{} & \multicolumn{3}{c}{\textit{Text $\longrightarrow$ Video}} \\
\multicolumn{2}{c|}{Word embeddings} & Aggregation & R@5$\uparrow$ & MdR$\downarrow$ & MnR$\downarrow$ \\
\hline
\multirow{4}{*}{GrOVLE} & \multirow{2}{*}{frozen} & maxpool & $\hspace{1.5em}31.8_{\!\pm\!0.4}$ & $\hspace{1.5em}14.7_{\!\pm\!0.5}$ & $\hspace{1.5em}63.1_{\!\pm\!1.3}$ \\
                                                & & LSTM & $\hspace{1.5em}36.4_{\!\pm\!0.8}$ & $\hspace{1.5em}10.3_{\!\pm\!0.9}$ & $\hspace{1.5em}44.2_{\!\pm\!0.1}$ \\
                        \cline{2-6}
                         & \multirow{2}{*}{finetuned} & maxpool & $\hspace{1.5em}34.6_{\!\pm\!0.1}$ & $\hspace{1.5em}12.0_{\!\pm\!0.0}$ & $\hspace{1.5em}52.3_{\!\pm\!0.8}$ \\
                         & & LSTM & $\hspace{1.5em}40.3_{\!\pm\!0.5}$ & $\hspace{1.5em}8.7_{\!\pm\!0.5}$ & $\hspace{1.5em}38.1_{\!\pm\!0.7}$ \\
\hline
\multirow{6}{*}{BERT} & \multirow{2}{*}{frozen} & maxpool & $\hspace{1.5em}39.4_{\!\pm\!0.8}$ & $\hspace{1.5em}9.7_{\!\pm\!0.5}$ & $\hspace{1.5em}46.5_{\!\pm\!0.2}$ \\
                         & & LSTM & $\hspace{1.5em}36.4_{\!\pm\!1.8}$ & $\hspace{1.5em}10.7_{\!\pm\!0.5}$ & $\hspace{1.5em}42.2_{\!\pm\!0.6}$ \\
                         \cline{2-6}
                         & \multirow{2}{*}{finetuned} & maxpool & $\hspace{1.5em}44.2_{\!\pm\!1.2}$ & $\hspace{1.5em}7.3_{\!\pm\!0.5}$ & $\hspace{1.5em}35.6_{\!\pm\!0.4}$ \\
                         & & LSTM & $\hspace{1.5em}40.1_{\!\pm\!1.0}$ & $\hspace{1.5em}8.7_{\!\pm\!0.5}$ & $\hspace{1.5em}37.4_{\!\pm\!0.5}$ \\
                         \cline{2-6}
                         & frozen & BERT-frozen & $\hspace{1.5em}17.1_{\!\pm\!0.2}$ & $\hspace{1.5em}34.7_{\!\pm\!1.2}$ & $\hspace{1.5em}98.8_{\!\pm\!0.8}$ \\
                         & finetuned & BERT-finetuned & $\hspace{1.5em}\textbf{54.0}_{\!\pm\!0.2}$ & $\hspace{1.5em}\textbf{4.0}_{\!\pm\!0.0}$ & $\hspace{1.5em}\textbf{26.7}_{\!\pm\!0.9}$ \\
\hline
\end{tabular}
\end{center}
\end{table}

We also analysed the impact of removing stop words from the captions in Table~\ref{table:remove_stop_words}.
In a zero-shot setting, i.e., trained on HowTo100M, evaluated on MSRVTT without finetuning, removing the stop words helps generalize, by bridging the domain gap---HowTo100M speech is very different from MSRVTT captions. This approach was adopted in~\cite{miech2019MIL-NCE}. However, we observe that when finetuning, it is better to keep all the words as they contribute to the semantics of the caption.

\noindent\textbf{Video encoder.}
We evaluated the influence of different architectures for computing video embeddings on the MSRVTT 1k-A test split.

\begin{table}[t]
  \caption{Ablation studies on the video encoder of our framework with MSRVTT.
  \textbf{(a) Influence of the architecture and input.} With max-pooled features as input, we compare our transformer architecture (MMT) with the variant not using an encoder (NONE) and the one with Collaborative Gating~\cite{liu2019use} (COLL). We also show that MMT can attend to all extracted features, as detailed in the text. %
  \textbf{(b) Importance of initializing $F_{agg}^n$ features.} We compare zero-vector initialisation, mean pooling and max pooling of the expert features.
  \textbf{(c) Influence of the size of the multi-modal transformer.} We compare different values for number-of-layers $\times$ number-of-attention-heads.}
  \begin{subtable}[t]{1.\linewidth}%
    \centering%
    \caption{Encoder architecture and input}
        \begin{tabular}{l | l | @{\hskip -0.35cm}c @{\hskip -0.35cm}c @{\hskip -0.35cm}c}
        \hline
        \multicolumn{1}{c}{} & \multicolumn{1}{c}{} & \multicolumn{3}{c}{\textit{Text $\longrightarrow$ Video}} \\
        Encoder & Input & R@5$\uparrow$ & MdR$\downarrow$ & MnR$\downarrow$ \\
        \hline
        NONE & max pool & $\hspace{1.5em}50.9_{\!\pm\!1.5}$ & $\hspace{1.5em}5.3_{\!\pm\!0.5}$ & $\hspace{1.5em}28.6_{\!\pm\!0.5}$ \\
        COLL & max pool & $\hspace{1.5em}51.3_{\!\pm\!0.8}$ & $\hspace{1.5em}5.0_{\!\pm\!0.0}$ & $\hspace{1.5em}29.5_{\!\pm\!1.8}$ \\
        MMT & max pool & $\hspace{1.5em}52.5_{\!\pm\!0.7}$ & $\hspace{1.5em}5.0_{\!\pm\!0.0}$ & $\hspace{1.5em}27.2_{\!\pm\!0.7}$ \\
        MMT & shuffled feats & $\hspace{1.5em}53.3_{\!\pm\!0.2}$ & $\hspace{1.5em}5.0_{\!\pm\!0.0}$ & $\hspace{1.5em}27.4_{\!\pm\!0.7}$ \\
        MMT & ordered feats & $\hspace{1.5em}\textbf{54.0}_{\!\pm\!0.2}$ & $\hspace{1.5em}\textbf{4.0}_{\!\pm\!0.0}$ & $\hspace{1.5em}\textbf{26.7}_{\!\pm\!0.9}$ \\
        \hline
        \end{tabular}
        \label{table:video_model-cross_modal}
  \end{subtable}\par\bigskip
  \begin{subtable}[t]{.5\linewidth}%
    \centering%
        \caption{$F_{agg}^n$ initialisation}
        \begin{tabular}{l | @{\hskip -0.35cm}c @{\hskip -0.35cm}c @{\hskip -0.35cm}c}
        \hline
        \multicolumn{1}{c}{} & \multicolumn{3}{c}{\textit{Text $\longrightarrow$ Video}} \\
        $F_{agg}^n$ init & R@5$\uparrow$ & MdR$\downarrow$ & MnR$\downarrow$ \\
        \hline
        zero & $\hspace{1.5em}50.2_{\!\pm\!0.9}$ & $\hspace{1.5em}5.7_{\!\pm\!0.5}$ & $\hspace{1.5em}28.5_{\!\pm\!1.3}$ \\
        mean pool & $\hspace{1.5em}\textbf{54.2}_{\!\pm\!0.3}$ & $\hspace{1.5em}5.0_{\!\pm\!0.0}$ & $\hspace{1.5em}27.1_{\!\pm\!0.9}$ \\
        max pool& $\hspace{1.5em}54.0_{\!\pm\!0.2}$ & $\hspace{1.5em}\textbf{4.0}_{\!\pm\!0.0}$ & $\hspace{1.5em}\textbf{26.7}_{\!\pm\!0.9}$ \\
        \hline
        \end{tabular}
        \label{table:video_model-initialisation}
  \end{subtable}%
  \begin{subtable}[t]{.5\linewidth}
    \centering
        \caption{Model size}
        \begin{tabular}{l | l | @{\hskip -0.35cm}c @{\hskip -0.35cm}c @{\hskip -0.35cm}c}
        \hline
        \multicolumn{1}{c}{} & \multicolumn{1}{c}{} & \multicolumn{3}{c}{\textit{Text $\longrightarrow$ Video}} \\
        Layers & Heads & R@5$\uparrow$ & MdR$\downarrow$ & MnR$\downarrow$ \\
        \hline
        2 & 2 & $\hspace{1.5em}53.2_{\!\pm\!0.4}$ & $\hspace{1.5em}5.0_{\!\pm\!0.0}$ & $\hspace{1.5em}\textbf{26.7}_{\!\pm\!0.4}$ \\
        4 & 4 & $\hspace{1.5em}\textbf{54.0}_{\!\pm\!0.2}$ & $\hspace{1.5em}\textbf{4.0}_{\!\pm\!0.0}$ & $\hspace{1.5em}\textbf{26.7}_{\!\pm\!0.9}$ \\
        8 & 8 & $\hspace{1.5em}53.9_{\!\pm\!0.3}$ & $\hspace{1.5em}4.7_{\!\pm\!0.5}$ & $\hspace{1.5em}\textbf{26.7}_{\!\pm\!0.7}$ \\
        \hline
        \end{tabular}
        \label{table:video_model-model_size}
  \end{subtable}
  \label{table:video_model-ablations}
\end{table}

In Table~\ref{table:video_model-cross_modal}, we evaluate variants of our encoder architecture and its input. 
Similar to~\cite{miech2018learning}, we experiment with directly computing the caption-video similarities on each max-pooled expert features, i.e., no video encoder (NONE in the table). We compare this with the collaborative gating architecture (COLL)~\cite{liu2019use} and our MMT variant using only the aggregated features as input. For the first two variants without MMT, we adopt the approach of~\cite{miech2018learning} to deal with missing modalities by re-weighting $w_i(c)$. 
We also show the superior performance of our multi-modal transformer in contextualising the different modality embeddings compared to the collaborative gating approach. We argue that our MMT is able to extract cross-modal information in a multi-stage architecture compared to collaborative gating, which is limited to modulating the input embeddings. 
Table~\ref{table:video_model-cross_modal} also highlights the advantage of providing MMT with \textbf{all} the extracted features, instead of only  aggregated ones. 
Temporally aggregating each expert's features ignores information about multiple events occurring in a same video (see the last three rows). 
As shown by the influence of ordered and randomly shuffled features on the performance, MMT has the capacity to make sense of the relative ordering of events in a video.

Table~\ref{table:video_model-initialisation} shows the importance of initialising the expert aggregation feature $F_{agg}^n$. Since the output of our video encoder is extracted from the ``agg'' columns, it is important to initialise them with an appropriate representation of the experts' features. The transformer being a residual network architecture, initializing $F_{agg}^n$ input embeddings with a zero vector leads to a low performance. Initializing with max pooling aggregation of each expert performs better than mean pooling.
Finally, we analyze the impact of the size of our multi-modal transformer model in Table~\ref{table:video_model-model_size}. A model with 4 layers and 4 attention heads outperforms both a smaller model (2 layers and 2 attention heads) and a larger model (8 layers and 8 attention heads).

\begin{figure}[t]
    \centering
    \includegraphics[width=.32\linewidth]{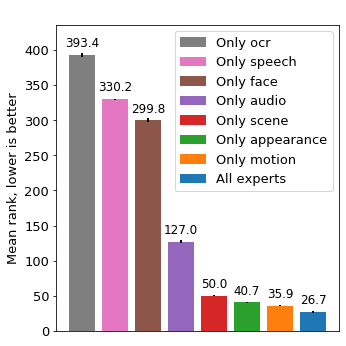}
    \includegraphics[width=.32\linewidth]{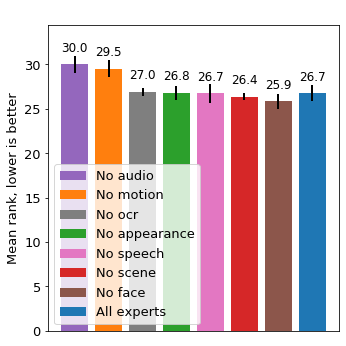}
    \includegraphics[width=.32\linewidth]{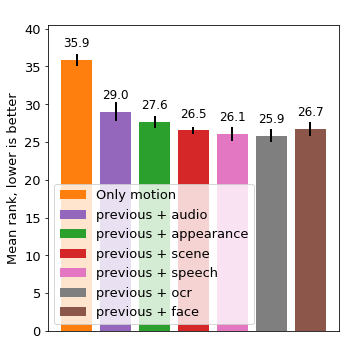}
    \caption{MSRVTT performance (mean rank; lower is better) after training from scratch, when using only one expert (left), when using all experts but one (middle), when gradually adding experts by greedy search (right).}
    \label{fig:experts_ablation}
\end{figure}

\noindent\textbf{Comparison of the different experts.} In Figure~\ref{fig:experts_ablation}, we show an ablation study when training our model on MSRVTT using only one expert (left), using all experts but one (middle), or gradually adding experts by greedy search (right). 
In the case of using only one expert, we note that the motion expert provides the best results. We attribute the poor performance of OCR, speech and face to the fact that they are absent from many videos, thus resulting in a zero vector input to our video encoder.
While the scene expert shows a decent performance, if used alone, it does not contribute when used along other experts, perhaps due to the semantics it encodes being captured already by other experts like appearance or motion. On the contrary, the audio expert alone does not provide a good performance, but it contributes the most when used in conjunction with the others, most likely due to the complementary cues it provides, compared to the other experts.

\noindent\textbf{Comparison to prior state of the art.}
We compare our method on three  datasets: MSRVTT (Table~\ref{table:MSRVTT_results}), ActivityNet (Table~\ref{table:ANet_results}) and LSMDC (Table~\ref{table:LSMDC_results}). While MSRVTT and LSMDC contain short video-caption pairs (average video duration of 13s for MSRVTT, one-sentence captions), ActivityNet contains much longer videos (several minutes) and each video is captioned with multiple sentences. We consider the concatenation of all these sentences as the caption. We show that our method obtains state-of-the-art results on all the three datasets. The gains obtained through MMT's long term temporal encoding are particularly noticeable on the long videos of ActivityNet. 

\begin{table}[h!]
\begin{center}
\caption{Retrieval performance on the MSRVTT dataset. 1k-A and 1k-B denote test sets of 1000 randomly sampled caption-video pairs used in~\cite{Yu2018JSFusion} and~\cite{miech2018learning} resp.}
\label{table:MSRVTT_results}
\scriptsize
\begin{tabular}{l | l| @{\hskip -0.35cm}c @{\hskip -0.35cm}c @{\hskip -0.35cm}c | @{\hskip -0.35cm}c @{\hskip -0.35cm}c @{\hskip -0.35cm}c}
\hline
\multicolumn{1}{c}{} & \multicolumn{1}{c}{} & \multicolumn{3}{c}{\textit{Text $\longrightarrow$ Video}} & \multicolumn{3}{c}{\textit{Video $\longrightarrow$ Text}} \\
Method & Split & R@5$\uparrow$ & MdR$\downarrow$ & MnR$\downarrow$ & R@5$\uparrow$ & MdR$\downarrow$ & MnR$\downarrow$ \\
\hline
Random baseline & 1k-A & 0.5 & 500.0 & 500.0 & 0.5 & 500.0 & 500.0 \\
JSFusion~\cite{Yu2018JSFusion} & 1k-A & 31.2 & 13 & - & - & - & - \\
HT~\cite{miech19howto100m} & 1k-A & 35.0 & 12 & - & - & - & - \\
CE~\cite{liu2019use} & 1k-A & $\hspace{1.5em}48.8_{\!\pm\!0.6}$ & $\hspace{1.5em}6.0_{\!\pm\!0.0}$ & $\hspace{1.5em}28.2_{\!\pm\!0.8}$ & $\hspace{1.5em}50.3_{\!\pm\!0.5}$ & $\hspace{1.5em}5.3_{\!\pm\!0.6}$ & $\hspace{1.5em}25.1_{\!\pm\!0.8}$ \\
Ours & 1k-A & $\hspace{1.5em}\textbf{54.0}_{\!\pm\!0.2}$ & $\hspace{1.5em}\textbf{4.0}_{\!\pm\!0.0}$ & $\hspace{1.5em}\textbf{26.7}_{\!\pm\!0.9}$ & $\hspace{1.5em}\textbf{56.0}_{\!\pm\!0.9}$ & $\hspace{1.5em}\textbf{4.0}_{\!\pm\!0.0}$ & $\hspace{1.5em}\textbf{23.6}_{\!\pm\!1.0}$ \\
\hline
HT-pretrained~\cite{miech19howto100m} & 1k-A & 40.2 & 9 & - & - & - & - \\
Ours-pretrained & 1k-A & $\hspace{1.5em}\textbf{57.1}_{\!\pm\!1.0}$ & $\hspace{1.5em}\textbf{4.0}_{\!\pm\!0.0}$ & $\hspace{1.5em}\textbf{24.0}_{\!\pm\!0.8}$ & $\hspace{1.5em}\textbf{57.5}_{\!\pm\!0.6}$ & $\hspace{1.5em}\textbf{3.7}_{\!\pm\!0.5}$ & $\hspace{1.5em}\textbf{21.3}_{\!\pm\!0.6}$ \\
\hline\hline
Random baseline & 1k-B & 0.5 & 500.0 & 500.0 & 0.5 & 500.0 & 500.0 \\
MEE~\cite{miech2018learning} & 1k-B & 37.9 & 10.0 & - & - & - & - \\
JPose~\cite{wray2019finegrained} & 1k-B & 38.1 & 9 & - & 41.3 & 8.7 & - \\
MEE-COCO~\cite{miech2018learning} & 1k-B & 39.2 & 9.0 & - & - & - & - \\
CE~\cite{liu2019use} & 1k-B & $\hspace{1.5em}46.0_{\!\pm\!0.4}$ & $\hspace{1.5em}7.0_{\!\pm\!0.0}$ & $\hspace{1.5em}35.3_{\!\pm\!1.1}$ & $\hspace{1.5em}46.0_{\!\pm\!0.5}$ & $\hspace{1.5em}6.5_{\!\pm\!0.5}$ & $\hspace{1.5em}30.6_{\!\pm\!1.2}$ \\
Ours & 1k-B & $\hspace{1.5em}\textbf{49.1}_{\!\pm\!0.4}$ & $\hspace{1.5em}\textbf{6.0}_{\!\pm\!0.0}$ & $\hspace{1.5em}\textbf{29.5}_{\!\pm\!1.6}$ & $\hspace{1.5em}\textbf{49.4}_{\!\pm\!0.4}$ & $\hspace{1.5em}\textbf{6.0}_{\!\pm\!0.0}$ & $\hspace{1.5em}\textbf{24.5}_{\!\pm\!1.8}$ \\
\hline
\end{tabular}
\end{center}
\end{table}

\begin{table}[h!]
\begin{center}
\caption{Retrieval performance on the ActivityNet dataset.}
\label{table:ANet_results}
\scriptsize
\begin{tabular}{l | @{\hskip -0.35cm}c @{\hskip -0.35cm}c @{\hskip -0.35cm}c | @{\hskip -0.35cm}c @{\hskip -0.35cm}c @{\hskip -0.35cm}c}
\hline
\multicolumn{1}{c}{} & \multicolumn{3}{c}{\textit{Text $\longrightarrow$ Video}} & \multicolumn{3}{c}{\textit{Video $\longrightarrow$ Text}} \\
Method & R@5$\uparrow$ & MdR$\downarrow$ & MnR$\downarrow$ & R@5$\uparrow$ & MdR$\downarrow$ & MnR$\downarrow$ \\
\hline
Random baseline & 0.1 & 2458.5 & 2458.5 & 0.1 & 2458.5 & 2458.5 \\
FSE~\cite{zhang2018HSE} & $\hspace{1.5em}44.8_{\!\pm\!0.4}$ & 7 & - & $\hspace{1.5em}43.1_{\!\pm\!1.1}$ & 7 & - \\
CE~\cite{liu2019use} & $\hspace{1.5em}47.7_{\!\pm\!0.6}$ & $\hspace{1.5em}6.0_{\!\pm\!0.0}$ & $\hspace{1.5em}23.1_{\!\pm\!0.5}$ & $\hspace{1.5em}46.6_{\!\pm\!0.7}$ & $\hspace{1.5em}6.0_{\!\pm\!0.0}$ & $\hspace{1.5em}24.4_{\!\pm\!0.5}$ \\
HSE~\cite{zhang2018HSE}& 49.3 & - & - & 48.1 & - & - \\
Ours & $\hspace{1.5em}\textbf{54.2}_{\!\pm\!1.0}$ & $\hspace{1.5em}\textbf{5.0}_{\!\pm\!0.0}$ & $\hspace{1.5em}\textbf{20.8}_{\!\pm\!0.4}$ & $\hspace{1.5em}\textbf{54.8}_{\!\pm\!0.4}$ & $\hspace{1.5em}\textbf{4.3}_{\!\pm\!0.5}$ & $\hspace{1.5em}\textbf{21.2}_{\!\pm\!0.5}$ \\
\hline
Ours-pretrained & $\hspace{1.5em}\textbf{61.4}_{\!\pm\!0.2}$ & $\hspace{1.5em}\textbf{3.3}_{\!\pm\!0.5}$ & $\hspace{1.5em}\textbf{16.0}_{\!\pm\!0.4}$ & $\hspace{1.5em}\textbf{61.1}_{\!\pm\!0.2}$ & $\hspace{1.5em}\textbf{4.0}_{\!\pm\!0.0}$ & $\hspace{1.5em}\textbf{17.1}_{\!\pm\!0.5}$ \\
\hline
\end{tabular}
\end{center}
\end{table}

\begin{table}[h!]
\begin{center}
\caption{Retrieval performance on the LSMDC dataset.}
\label{table:LSMDC_results}
\scriptsize
\begin{tabular}{l | @{\hskip -0.35cm}c @{\hskip -0.35cm}c @{\hskip -0.35cm}c | @{\hskip -0.35cm}c @{\hskip -0.35cm}c @{\hskip -0.35cm}c}
\hline
\multicolumn{1}{c}{} & \multicolumn{3}{c}{\textit{Text $\longrightarrow$ Video}} & \multicolumn{3}{c}{\textit{Video $\longrightarrow$ Text}} \\
Method & R@5$\uparrow$ & MdR$\downarrow$ & MnR$\downarrow$ & R@5$\uparrow$ & MdR$\downarrow$ & MnR$\downarrow$ \\
\hline
Random baseline & 0.5 & 500.0 & 500.0 & 0.5 & 500.0 & 500.0 \\
CT-SAN~\cite{Yu2016CT-SAN} & 16.3 & 46 & - & - & - & - \\
JSFusion~\cite{Yu2018JSFusion} & 21.2 & 36 & - & - & - & - \\
CCA~\cite{Klein2015CCA} (rep. by~\cite{miech2018learning}) & 21.7 & 33 & - & - & - & - \\
MEE~\cite{miech2018learning} & 25.1 & 27 & - & - & - & - \\
MEE-COCO~\cite{miech2018learning} & 25.6 & 27 & - & - & - & - \\
CE~\cite{liu2019use} & $\hspace{1.5em}26.9_{\!\pm\!1.1}$ & $\hspace{1.5em}25.3_{\!\pm\!3.1}$ & - & - & - & - \\
Ours & $\hspace{1.5em}\textbf{29.2}_{\!\pm\!0.8}$ & $\hspace{1.5em}\textbf{21.0}_{\!\pm\!1.4}$ & $\hspace{1.5em}\textbf{76.3}_{\!\pm\!1.9}$ & $\hspace{1.5em}\textbf{29.3}_{\!\pm\!1.1}$ & $\hspace{1.5em}\textbf{22.5}_{\!\pm\!0.4}$ & $\hspace{1.5em}\textbf{77.1}_{\!\pm\!2.6}$ \\
\hline
Ours-pretrained & $\hspace{1.5em}\textbf{29.9}_{\!\pm\!0.7}$ & $\hspace{1.5em}\textbf{19.3}_{\!\pm\!0.2}$ & $\hspace{1.5em}\textbf{75.0}_{\!\pm\!1.2}$ & $\hspace{1.5em}\textbf{28.6}_{\!\pm\!0.3}$ & $\hspace{1.5em}\textbf{20.0}_{\!\pm\!0.0}$ & $\hspace{1.5em}\textbf{76.0}_{\!\pm\!0.8}$ \\
\hline
\end{tabular}
\end{center}
\end{table}

\section{Summary}
We introduced multi-modal transformer, a transformer-based architecture capable of attending multiple features extracted at different moments, and from different modalities in video. This leverages both temporal and cross-modal cues, which are crucial for accurate video representation. We incorporate this video encoder along with a caption encoder in a cross-modal framework to perform caption-video matching and obtain state-of-the-art results for video retrieval. As future work, we would like to improve temporal encoding for video and text.

\paragraph{Acknowledgments.}
We thank the authors of~\cite{liu2019use} for sharing their codebase and features, and Samuel Albanie, in particular, for his help with implementation details. This work was supported in part by the ANR project AVENUE.

\bibliographystyle{splncs04}
\bibliography{lib.bib}

\newpage
\appendix
\section{Supplementary material}

\subsection{Model complexity}

\paragraph{Number of parameters.} As shown below, using multiple modalities does not impact the number of parameters significantly. Interestingly, majority of the parameters correspond to the BERT caption encoding module. We also note that the difference in the video encoder comes from the projections. The number of parameters of a transformer encoder is independent of the number of input embeddings, as are the parameters of a CNN from the image size.

Our cross-modal architecture using 7 modalities has: 
133.3M parameters, including caption encoder: 112.9M, video encoder: 20.4M (Projections: 3.3M, MMT: 17.1M).
Our cross-modal architecture using 2 modalities has:
127.3M parameters, including caption encoder: 109.6M (decrease compared to 7 modalities due to using less gated embedding modules), video encoder: 17.7M (Projections: 0.6M, MMT: 17.1M).

\paragraph{Training and inference times.} Training our full cross-modal architecture from scratch on MSRVTT takes about 4 hours on a single V100 16GB GPU.

If we replace our multi-modal transformer by collaborative gating~\cite{liu2019use}, we reduce the number of parameters from 133.3M to 123.9M. However, the gain in inference time is minimal, from 1.1s to 0.8s, and is negligible compared to feature extraction, as detailed below.

Inference time for 1k videos and 1k text queries from MSRVTT on a single V100 GPU is as follows: approximately 3000s to extract features of 7 experts on 1k videos (480s just for S3D motion features), 1.1s to process videos with MMT, 0.9s to process 1k captions with BERT+gated embedding modules, 0.05s to compute similarities and rank the video candidates for the 1k queries.

\subsection{Results on additional metrics}
Here, we report our results for the additional metrics R@1, R@10, R@50. Table~\ref{table:MSRVTT_results_supp} complements the results reported for the MSRVTT~\cite{xu2016msrvtt} dataset in 
Table~\ref{table:MSRVTT_results}
of the main paper. Similarly, Table~\ref{table:ANet_results_supp} and Table~\ref{table:LSMDC_results_supp} report the additional evaluations  for  
Table~\ref{table:ANet_results} and Table~\ref{table:LSMDC_results} 
of the main paper on ActivityNet~\cite{krishna2017activitynet} and LSMDC~\cite{Rohrbach2015LSMDC} datasets respectively. 
We observe that the results on these additional metrics are in line with the conclusions of the main paper.

\begin{table}[h!]
\begin{center}
\caption{Retrieval performance on the MSRVTT dataset. 1k-A and 1k-B denote test sets of 1000 randomly sampled caption-video pairs used in~\cite{Yu2018JSFusion} and~\cite{miech2018learning} resp.}
\label{table:MSRVTT_results_supp}
\scriptsize
\scalebox{0.90}{
\begin{tabular}{l | l| @{\hskip -0.35cm}c @{\hskip -0.35cm}c @{\hskip -0.35cm}c @{\hskip -0.35cm}c @{\hskip -0.35cm}c | @{\hskip -0.35cm}c @{\hskip -0.35cm}c @{\hskip -0.35cm}c @{\hskip -0.35cm}c @{\hskip -0.35cm}c}
\hline
\multicolumn{1}{c}{} & \multicolumn{1}{c}{} & \multicolumn{5}{c}{\textit{Text $\longrightarrow$ Video}} & \multicolumn{5}{c}{\textit{Video $\longrightarrow$ Text}} \\
Method & Split & R@1$\uparrow$ & R@5$\uparrow$ & R@10$\uparrow$ & MdR$\downarrow$ & MnR$\downarrow$ & R@1$\uparrow$ & R@5$\uparrow$ & R@10$\uparrow$ & MdR$\downarrow$ & MnR$\downarrow$ \\
\hline
Random baseline & 1k-A & 0.1 & 0.5 & 1.0 & 500.0 & 500.0 & 0.1 & 0.5 & 1.0 & 500.0 & 500.0 \\
JSFusion~\cite{Yu2018JSFusion} & 1k-A & 10.2 & 31.2 & 43.2 & 13 & - & - & - & - & - & - \\
HT~\cite{miech19howto100m} & 1k-A & 12.1 & 35.0 & 48.0 & 12 & - & - & - & - & - & - \\
CE~\cite{liu2019use} & 1k-A & $\hspace{1.5em}20.9_{\!\pm\!1.2}$ & $\hspace{1.5em}48.8_{\!\pm\!0.6}$ & $\hspace{1.5em}62.4_{\!\pm\!0.8}$ & $\hspace{1.5em}6.0_{\!\pm\!0.0}$ & $\hspace{1.5em}28.2_{\!\pm\!0.8}$ & $\hspace{1.5em}20.6_{\!\pm\!0.6}$ & $\hspace{1.5em}50.3_{\!\pm\!0.5}$ & $\hspace{1.5em}64.0_{\!\pm\!0.2}$ & $\hspace{1.5em}5.3_{\!\pm\!0.6}$ & $\hspace{1.5em}25.1_{\!\pm\!0.8}$ \\
Ours & 1k-A & $\hspace{1.5em}\textbf{24.6}_{\!\pm\!0.4}$ & $\hspace{1.5em}\textbf{54.0}_{\!\pm\!0.2}$ & $\hspace{1.5em}\textbf{67.1}_{\!\pm\!0.5}$ & $\hspace{1.5em}\textbf{4.0}_{\!\pm\!0.0}$ & $\hspace{1.5em}\textbf{26.7}_{\!\pm\!0.9}$ & $\hspace{1.5em}\textbf{24.4}_{\!\pm\!0.5}$ & $\hspace{1.5em}\textbf{56.0}_{\!\pm\!0.9}$ & $\hspace{1.5em}\textbf{67.8}_{\!\pm\!0.3}$ & $\hspace{1.5em}\textbf{4.0}_{\!\pm\!0.0}$ & $\hspace{1.5em}\textbf{23.6}_{\!\pm\!1.0}$ \\
\hline
HT-pretrained~\cite{miech19howto100m} & 1k-A & 14.9 & 40.2 & 52.8 & 9 & - & - & - & - & - & - \\
Ours-pretrained & 1k-A & $\hspace{1.5em}\textbf{26.6}_{\!\pm\!1.0}$ & $\hspace{1.5em}\textbf{57.1}_{\!\pm\!1.0}$ & $\hspace{1.5em}\textbf{69.6}_{\!\pm\!0.2}$ & $\hspace{1.5em}\textbf{4.0}_{\!\pm\!0.0}$ & $\hspace{1.5em}\textbf{24.0}_{\!\pm\!0.8}$ & $\hspace{1.5em}\textbf{27.0}_{\!\pm\!0.6}$ & $\hspace{1.5em}\textbf{57.5}_{\!\pm\!0.6}$ & $\hspace{1.5em}\textbf{69.7}_{\!\pm\!0.8}$ & $\hspace{1.5em}\textbf{3.7}_{\!\pm\!0.5}$ & $\hspace{1.5em}\textbf{21.3}_{\!\pm\!0.6}$ \\
\hline\hline
Random baseline & 1k-B & 0.1 & 0.5 & 1.0 & 500.0 & 500.0 & 0.1 & 0.5 & 1.0 & 500.0 & 500.0 \\
MEE~\cite{miech2018learning} & 1k-B & 13.6 & 37.9 & 51.0 & 10.0 & - & - & - & - & - & - \\
JPose~\cite{wray2019finegrained} & 1k-B & 14.3 & 38.1 & 53.0 & 9 & - & 16.4 & 41.3 & 54.4 & 8.7 & - \\
MEE-COCO~\cite{miech2018learning} & 1k-B & 14.2 & 39.2 & 53.8 & 9.0 & - & - & - & - & - & - \\
CE~\cite{liu2019use} & 1k-B & $\hspace{1.5em}18.2_{\!\pm\!0.7}$ & $\hspace{1.5em}46.0_{\!\pm\!0.4}$ & $\hspace{1.5em}60.7_{\!\pm\!0.2}$ & $\hspace{1.5em}7.0_{\!\pm\!0.0}$ & $\hspace{1.5em}35.3_{\!\pm\!1.1}$ & $\hspace{1.5em}18.0_{\!\pm\!0.8}$ & $\hspace{1.5em}46.0_{\!\pm\!0.5}$ & $\hspace{1.5em}60.3_{\!\pm\!0.5}$ & $\hspace{1.5em}6.5_{\!\pm\!0.5}$ & $\hspace{1.5em}30.6_{\!\pm\!1.2}$ \\
Ours & 1k-B & $\hspace{1.5em}\textbf{20.3}_{\!\pm\!0.5}$ & $\hspace{1.5em}\textbf{49.1}_{\!\pm\!0.4}$ & $\hspace{1.5em}\textbf{63.9}_{\!\pm\!0.5}$ & $\hspace{1.5em}\textbf{6.0}_{\!\pm\!0.0}$ & $\hspace{1.5em}\textbf{29.5}_{\!\pm\!1.6}$ & $\hspace{1.5em}\textbf{21.1}_{\!\pm\!0.4}$ & $\hspace{1.5em}\textbf{49.4}_{\!\pm\!0.4}$ & $\hspace{1.5em}\textbf{63.2}_{\!\pm\!0.4}$ & $\hspace{1.5em}\textbf{6.0}_{\!\pm\!0.0}$ & $\hspace{1.5em}\textbf{24.5}_{\!\pm\!1.8}$ \\
\hline
\end{tabular}
}
\end{center}
\vspace{-4mm}
\end{table}

\begin{table}[h!]
\begin{center}
\caption{Retrieval performance on the ActivityNet dataset.}
\label{table:ANet_results_supp}
\scriptsize
\scalebox{0.90}{
\begin{tabular}{l | @{\hskip -0.35cm}c @{\hskip -0.35cm}c @{\hskip -0.35cm}c @{\hskip -0.35cm}c @{\hskip -0.35cm}c | @{\hskip -0.35cm}c @{\hskip -0.35cm}c @{\hskip -0.35cm}c @{\hskip -0.35cm}c @{\hskip -0.35cm}c}
\hline
\multicolumn{1}{c}{} & \multicolumn{5}{c}{\textit{Text $\longrightarrow$ Video}} & \multicolumn{5}{c}{\textit{Video $\longrightarrow$ Text}} \\
Method & R@1$\uparrow$ & R@5$\uparrow$ & R@50$\uparrow$ & MdR$\downarrow$ & MnR$\downarrow$ & R@1$\uparrow$ & R@5$\uparrow$ & R@50$\uparrow$ & MdR$\downarrow$ & MnR$\downarrow$ \\
\hline
Random baseline & 0.02 & 0.1 & 1.02 & 2458.5 & 2458.5 & 0.02 & 0.1 & 1.02 & 2458.5 & 2458.5 \\
FSE~\cite{zhang2018HSE} & $\hspace{1.5em}18.2_{\!\pm\!0.2}$ & $\hspace{1.5em}44.8_{\!\pm\!0.4}$ & $\hspace{1.5em}89.1_{\!\pm\!0.3}$ & 7 & - & $\hspace{1.5em}16.7_{\!\pm\!0.8}$ & $\hspace{1.5em}43.1_{\!\pm\!1.1}$ & $\hspace{1.5em}88.4_{\!\pm\!0.3}$ & 7 & - \\
CE~\cite{liu2019use} & $\hspace{1.5em}18.2_{\!\pm\!0.3}$ & $\hspace{1.5em}47.7_{\!\pm\!0.6}$ & $\hspace{1.5em}91.4_{\!\pm\!0.4}$ & $\hspace{1.5em}6.0_{\!\pm\!0.0}$ & $\hspace{1.5em}23.1_{\!\pm\!0.5}$ & $\hspace{1.5em}17.7_{\!\pm\!0.6}$ & $\hspace{1.5em}46.6_{\!\pm\!0.7}$ & $\hspace{1.5em}90.9_{\!\pm\!0.2}$ & $\hspace{1.5em}6.0_{\!\pm\!0.0}$ & $\hspace{1.5em}24.4_{\!\pm\!0.5}$ \\
HSE~\cite{zhang2018HSE}& 20.5 & 49.3 & - & - & - & 18.7 & 48.1 & - & - & - \\
Ours & $\hspace{1.5em}\textbf{22.7}_{\!\pm\!0.2}$ & $\hspace{1.5em}\textbf{54.2}_{\!\pm\!1.0}$ & $\hspace{1.5em}\textbf{93.2}_{\!\pm\!0.4}$ & $\hspace{1.5em}\textbf{5.0}_{\!\pm\!0.0}$ & $\hspace{1.5em}\textbf{20.8}_{\!\pm\!0.4}$ & $\hspace{1.5em}\textbf{22.9}_{\!\pm\!0.8}$ & $\hspace{1.5em}\textbf{54.8}_{\!\pm\!0.4}$ & $\hspace{1.5em}\textbf{93.1}_{\!\pm\!0.2}$ & $\hspace{1.5em}\textbf{4.3}_{\!\pm\!0.5}$ & $\hspace{1.5em}\textbf{21.2}_{\!\pm\!0.5}$ \\
\hline
Ours-pretrained & $\hspace{1.5em}\textbf{28.7}_{\!\pm\!0.2}$ & $\hspace{1.5em}\textbf{61.4}_{\!\pm\!0.2}$ & $\hspace{1.5em}\textbf{94.5}_{\!\pm\!0.0}$ & $\hspace{1.5em}\textbf{3.3}_{\!\pm\!0.5}$ & $\hspace{1.5em}\textbf{16.0}_{\!\pm\!0.4}$ & $\hspace{1.5em}\textbf{28.9}_{\!\pm\!0.2}$ & $\hspace{1.5em}\textbf{61.1}_{\!\pm\!0.2}$ & $\hspace{1.5em}\textbf{94.3}_{\!\pm\!0.4}$ & $\hspace{1.5em}\textbf{4.0}_{\!\pm\!0.0}$ & $\hspace{1.5em}\textbf{17.1}_{\!\pm\!0.5}$ \\
\hline
\end{tabular}
}
\end{center}
\vspace{-4mm}
\end{table}

\begin{table}[h!]
\begin{center}
\caption{Retrieval performance on the LSMDC dataset.}
\label{table:LSMDC_results_supp}
\scriptsize
\scalebox{0.90}{
\begin{tabular}{l | @{\hskip -0.35cm}c @{\hskip -0.35cm}c @{\hskip -0.35cm}c @{\hskip -0.35cm}c @{\hskip -0.35cm}c | @{\hskip -0.35cm}c @{\hskip -0.35cm}c @{\hskip -0.35cm}c @{\hskip -0.35cm}c @{\hskip -0.35cm}c}
\hline
\multicolumn{1}{c}{} & \multicolumn{5}{c}{\textit{Text $\longrightarrow$ Video}} & \multicolumn{5}{c}{\textit{Video $\longrightarrow$ Text}} \\
Method & R@1$\uparrow$ & R@5$\uparrow$ & R@10$\uparrow$ & MdR$\downarrow$ & MnR$\downarrow$ & R@1$\uparrow$ & R@5$\uparrow$ & R@10$\uparrow$ & MdR$\downarrow$ & MnR$\downarrow$ \\
\hline
Random baseline & 0.1 & 0.5 & 1.0 & 500.0 & 500.0 & 0.1 & 0.5 & 1.0 & 500.0 & 500.0 \\
CT-SAN~\cite{Yu2016CT-SAN} & 5.1 & 16.3 & 25.2 & 46 & - & - & - & - & - & - \\
JSFusion~\cite{Yu2018JSFusion} & 9.1 & 21.2 & 34.1 & 36 & - & - & - & - & - & - \\
CCA~\cite{Klein2015CCA} (rep. by~\cite{miech2018learning}) & 7.5 & 21.7 & 31.0 & 33 & - & - & - & - & - & - \\
MEE~\cite{miech2018learning} & 9.3 & 25.1 & 33.4 & 27 & - & - & - & - & - & - \\
MEE-COCO~\cite{miech2018learning} & 10.1 & 25.6 & 34.6 & 27 & - & - & - & - & - & - \\
CE~\cite{liu2019use} & $\hspace{1.5em}11.2_{\!\pm\!0.4}$ & $\hspace{1.5em}26.9_{\!\pm\!1.1}$ & $\hspace{1.5em}34.8_{\!\pm\!2.0}$ & $\hspace{1.5em}25.3_{\!\pm\!3.1}$ & - & - & - & - & - & - \\
Ours & $\hspace{1.5em}\textbf{13.2}_{\!\pm\!0.4}$ & $\hspace{1.5em}\textbf{29.2}_{\!\pm\!0.8}$ & $\hspace{1.5em}\textbf{38.8}_{\!\pm\!0.9}$ & $\hspace{1.5em}\textbf{21.0}_{\!\pm\!1.4}$ & $\hspace{1.5em}\textbf{76.3}_{\!\pm\!1.9}$ & $\hspace{1.5em}\textbf{12.1}_{\!\pm\!0.1}$ & $\hspace{1.5em}\textbf{29.3}_{\!\pm\!1.1}$ & $\hspace{1.5em}\textbf{37.9}_{\!\pm\!1.1}$ & $\hspace{1.5em}\textbf{22.5}_{\!\pm\!0.4}$ & $\hspace{1.5em}\textbf{77.1}_{\!\pm\!2.6}$ \\
\hline
Ours-pretrained & $\hspace{1.5em}\textbf{12.9}_{\!\pm\!0.1}$ & $\hspace{1.5em}\textbf{29.9}_{\!\pm\!0.7}$ & $\hspace{1.5em}\textbf{40.1}_{\!\pm\!0.8}$ & $\hspace{1.5em}\textbf{19.3}_{\!\pm\!0.2}$ & $\hspace{1.5em}\textbf{75.0}_{\!\pm\!1.2}$ & $\hspace{1.5em}\textbf{12.3}_{\!\pm\!0.2}$ & $\hspace{1.5em}\textbf{28.6}_{\!\pm\!0.3}$ & $\hspace{1.5em}\textbf{38.9}_{\!\pm\!0.8}$ & $\hspace{1.5em}\textbf{20.0}_{\!\pm\!0.0}$ & $\hspace{1.5em}\textbf{76.0}_{\!\pm\!0.8}$ \\
\hline
\end{tabular}
}
\end{center}
\vspace{-4mm}
\end{table}

\end{document}